\pdfoutput=1

\documentclass[11pt]{article}
\usepackage[dvipdfmx]{graphicx}
\usepackage{acl}
\usepackage{booktabs}
\usepackage{times}
\usepackage{latexsym}
\usepackage{natbib}
\usepackage{enumitem}
\usepackage[utf8]{inputenc}
\usepackage{amsmath}
\usepackage[T1]{fontenc}
\usepackage{CJKutf8}
\usepackage[utf8]{inputenc}
\usepackage{microtype}

\usepackage{inconsolata}
\usepackage{tabularx}
\usepackage{multirow}
\usepackage{multicol}
\usepackage{booktabs}
\usepackage{siunitx}
\title{CT-Eval: Benchmarking Chinese Text-to-Table Performance in Large Language Models}

\author{
Haoxiang Shi\textsuperscript{1},
Jiaan Wang,
Jiarong Xu\textsuperscript{2},
Cen Wang\textsuperscript{3} and  Tetsuya Sakai\textsuperscript{1} \\
\textsuperscript{1} Waseda University, Tokyo, Japan\ \  \textsuperscript{2} Fudan University \\
\textsuperscript{3} KDDI Research Inc., Japan\ \ \\
\small \texttt{hollis.shi@toki.waseda.jp}, \texttt{tetsuya@waseda.jp}, \texttt{xce-wang@kddi.com}, 
}

\begin{document}

\maketitle
\begin{abstract}

Text-to-Table aims to generate structured tables to convey the key information from unstructured documents.
Existing text-to-table datasets are typically oriented English, limiting the research in non-English languages.
Meanwhile, the emergence of large language models (LLMs) has shown great success as general task solvers in multi-lingual settings (\emph{e.g.}, \texttt{ChatGPT}), theoretically enabling text-to-table in other languages.
In this paper, we propose a Chinese text-to-table dataset, CT-Eval, to benchmark LLMs on this task.
Our preliminary analysis of English text-to-table datasets highlight two key factors for dataset construction: data diversity and data hallucination.
Inspired by this, the CT-Eval dataset selects a popular Chinese multidisciplinary online encyclopedia as the source and covers 28 domains to ensure data diversity.
To minimize data hallucination, we first train an LLM to judge and filter out the task samples with hallucination, then employ human annotators to clean the hallucinations in the validation and testing sets.
After this process, CT-Eval contains 88.6K task samples.

Using CT-Eval, we evaluate the performance of open-source and closed-source LLMs.
Our results reveal that zero-shot LLMs (including \texttt{GPT-4}) still have a significant performance gap compared with human judgment.
Furthermore, after fine-tuning, open-source LLMs can significantly improve their text-to-table ability, outperforming \texttt{GPT-4} by a large margin.
In short, CT-Eval not only helps researchers evaluate and quickly understand the Chinese text-to-table ability of existing LLMs but also serves as a valuable resource to significantly improve the text-to-table performance of LLMs.\footnote{
Codes and data will be publicly available once accepted.
}

\end{abstract}
\section{Introduction}

Information extraction (IE) aims to identify and extract structured information from unstructured text. Basic IE sub-tasks, such as, named entity recognition and entity linking, generally operate at the sentence level, which ignore models' ability to understand the document-level meaning.
In contrast, Text-to-Table, an emerging sub-task of IE, requires models to understand information within a given document and then generate structured tables.
Despite great success in the domain~\citet{wu2021text,li2023sequence}, previous studies are oriented to English, limiting the research in other languages.

Recently, large language models (LLMs) have demonstrated powerful performance across various NLP tasks~\cite{zhao2023survey,wang2023chatgpt,wang2023zero,wang2023cross}.
Some studies~\cite{gonzalez2023yes,gao2023exploring} have also utilized LLMs for several IE sub-tasks, and found that compared to supervised baselines, the performance of LLMs is sub-optimal.
However, \emph{their performance of LLMs on Text-to-Table remains largely unexplored}.
Additionally, the rapid advancements in LLMs have facilitated their widespread adoption in multi-lingual settings, enabling language modeling abilities to be shared across different languages.
Consequently, \emph{it is theoretically possible to feasible to employ LLMs for text-to-table in other languages, an area that has yet to be thoroughly investigated.}

Motivated by the aforementioned considerations, we decide to benchmark LLMs on Chinese text-to-table.
A heuristic approach involves translating existing English datasets into Chinese for subsequent performance evaluation.
There are four datasets widely used in text-to-table:
E2E~\cite{novikova2017e2e} is a restaurant domain dataset encompassing 51.5K samples about restaurant information like names, addresses, rate scores, etc.
Rotowire~\cite{wiseman2017challenges} is a sports domain dataset derived from NBA basketball games with 4.9K samples about NBA team information.
WikiBio~\cite{lebret2016neural}, compiled from Wikipedia, containing 72.6K documents and corresponding structured tables in the biography domain.
WikiTableText~\cite{bao2018table} is also derived from Wikipedia, and involves 13.3K multidisciplinary samples spanning fields like finance and politics.
However, according to our preliminary analysis, these datasets also suffer from the issues of \emph{less diversity} or \emph{high hallucination}, making them unsuitable to benchmark LLMs.
Specifically,
(1) E2E, Rotowire, and WikiBio exhibit a lack of diversity as they focus on a single domain, violating the core principle of instruction tuning in LLMs, that is, diversity.
(2) Although WikiTableText incorporates multiple domains for diversity, our analysis (\S~\ref{sec:Dataset Statics}) indicates that 18.83\% of samples exhibit hallucination in the golden tables, that is, containing additional information beyond the provided documents. This arises from treating Wikipedia infoboxes as golden tables, created collaboratively by online users and potentially containing additional basic information.

In this paper, we propose the \textbf{C}hinese \textbf{T}ext-to-Table \textbf{Eval}uation (CT-Eval) dataset, which is constructed through three steps to ensure data diversity and minimize hallucination.
To ensure diversity, the first step involves collecting multidisciplinary document-table pairs. We choose the Baidu Baike as the data source, which is a popular Chinese multidisciplinary online encyclopedia. Each page in this source contains text and an infobox summarizing the corresponding key information.
Second, to minimize data hallucination, we train an LLM, as a hallucination judger, to filter out task samples with hallucination in their golden tables (infoboxes).
Finally, we obtain 88.6K samples with an average length of 911.46 Chinese characters. We split them into 86.6K, 1K and 1K for training, validation and testing.
For validation and testing samples, human annotators further clean data hallucination in the golden tables to ensure evaluation reliability.

Based on the proposed CT-Eval, we benchmark various mainstream LLMs in both zero-shot (for both open- and closed-source LLMs) and fine-tuning (only for open-source LLMs) scenarios.
Our experiments reveal that (1) \texttt{GPT-4} achieves the best zero-shot performance among all LLMs. However, its performance remains a discernible disparity compared to human judgment.
(2) After fine-tuning on the training set of CT-Eval, all open-source LLMs demonstrate a significant performance improvement, outperforming zero-shot \texttt{GPT-4} by a large margin, indicating the effectiveness of CT-Eval.
In-depth analyses of LLM-generated tables reveal the persistence of hallucination issues in both zero-shot and fine-tuned LLMs, highlighting a challenge in using LLMs as text-to-table systems.
Future work could not only evaluate LLMs' performance on text-to-table via our CT-Eval benchmark dataset, but also leverage its training data to improve LLMs' text-to-table ability via fine-tuning.

\section{Related Work}

\subsection{Text-to-table Tasks}

\citet{wu2021text} pioneer the text-to-table task. Given the absence of a dedicated text-to-table dataset, \citet{wu2021text} repurpose existing table-to-text datasets, \emph{i.e.}, WikiBio~\cite{lebret2016neural}, E2E~\cite{novikova2017e2e}, Rotowire~\cite{wiseman2017challenges} and WikiTableText~\cite{bao2018table}, for text-to-table tasks by reversing their input-output pairs.
They fine-tune BART~\cite{lewis2019bart} to perform text-to-table in a sequence-to-sequence manner, and find that the fine-tuned BART outperforms the pipeline baselines using relation extraction and named entity extraction.
STable~\cite{pietruszka2022stable} employs two pre-trained language models (PLMs) (T5~\cite{raffel2020exploring} and TILT~\cite{powalski2021going}) for text-to-table, and designs a permutation-based decoder to enhance the PLMs' table generation ability.
Subsequently, \citet{li2023sequence} find that the predefined row order in golden tables introduces bias into text-to-table models.
Consequently, they train table header and table body generators separately to produce final tables.
While these studies achieve notable success, they primarily explored text-to-table performance before the LLM era.
In addition, their evaluation datasets generally focus on a single domain, and adapt from the table-to-text datasets, resulting in hallucination issues.
Thus, they are unsuitable for benchmarking LLMs in text-to-table.

\subsection{Large Language Models}

The advent of advanced LLMs ,\emph{e.g.}, \texttt{ChatGPT}~\cite{ChatGPT}, \texttt{GPT-4}~\cite{achiam2023gpt}, marks a pivotal moment that propels the field of NLP into a boom phase.
\citet{zhong2023can} show that LLMs can achieve decent performance on benchmarks like GLUE~\cite{wang2018glue}, which spans eight representative NLP understanding tasks.
Concurrently, several studies have scrutinized the performance of LLMs across various IE tasks. For example, \citet{gao2023exploring} test the capability of \texttt{ChatGPT} in event extraction.
Similarly, \citet{gonzalez2023yes} employ ChatGPT on historical entity recognition.
However, the results of these IE sub-tasks consistently show that LLMs underperform compared to state-of-the-art supervised approaches.
The text-to-table task we focused on is more complex than the previous basic IE sub-tasks, however, it remains an unexplored area for evaluation on LLMs.

\section{CT-Eval}
In this section, we first discuss the data source for building CT-Eval (\S~\ref{sec:Data Source}). Then, we give the details of how to control the hallucination in the preliminary collected data, including LLM hallucination judger (\S~\ref{sec:Primary Dataset Labeling}) and human cleaning processes (\S~\ref{sec:Hallucination Elimination}).
Finally, we formulate the text-to-table task (\S~\ref{sec:Task definition}) and provide the details of data statistics (\S~\ref{sec:Dataset Statics}).

\subsection{Data Source}\label{sec:Data Source}
Following the success of WikiTableText~\cite{bao2018table}, we also choose a multidisciplinary online encyclopedia as the data source to ensure data diversity.
After carefully comparing existing Chinese online encyclopedias, we choose Baidu Baike\footnote{\url{https://baike.baidu.com/}}, which is one of the Chinese encyclopedias with the most entries in the world.

We obtain the Baidu Baike data from the dumps provided by~\citet{xu2017cn}. The data contains over 9M entity pages, each of which includes an infobox and the corresponding textual description, forming a text-to-table sample.
Utilizing this data, we implement the following rule-based strategy for preliminary data cleaning:
(1) Each page must contain at least one infobox; otherwise, the golden table is missing.
(2) The number of tabular cells in the infobox should exceed three to ensure validity.
(3) The length of the textual description should surpass 200 tokens.
After that, 200K document-table pairs are remaining for further processing.

\subsection{LLM Data Cleaning}\label{sec:Primary Dataset Labeling}
After the initial cleaning process, the hallucination issue persists due to Baidu Baike's submission rules,  wherein most of the page text and infoboxes are edited and maintained by individuals. Thus, the contents in the infoboxes may not always align precisely with the textual documents.
For instance, some infoboxes may include additional knowledge unrelated to the text, potentially misleading the model from learning the text-to-table task.

\begin{figure}[t!]
\centering
\includegraphics[scale=0.5]{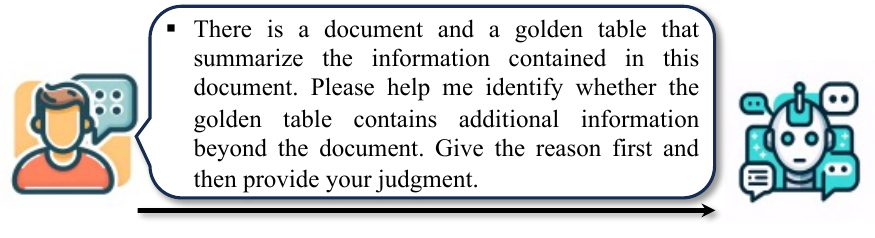}
\caption{Illustration of judgment prompt.}
\label{judgementprompt}
\end{figure}

To control the hallucination in the task samples, we decide to employ an LLM as a hallucination judger to filter out samples exhibiting hallucination.
While using GPT-4 directly as the LLM hallucination judger is a straightforward approach, utilizing official APIs can be costly.
Therefore, we first randomly select 5K samples from the data before cleaning.
Then, we use GPT-4 to assess whether the golden tables contain additional information.
The judgment prompt is illustrated in Figure~\ref{judgementprompt}, where the LLM is tasked with evaluating hallucination in a chain-of-thought manner.
Next, we use the GPT-4 judgment results to train an open-source LLM, employing the trained model to evaluate the remaining samples.
Samples containing hallucinations are discarded.
Given the capacity for understanding lengthy documents of existing Chinese LLMs~\cite{bai2023longbench}, we select ChatGLM3-6B-32k\footnote{\url{https://huggingface.co/THUDM/chatglm3-6b-32k}} as the open-source LLM hallucination judger. 
Finally, there are 88.6K samples after the data cleaning by the LLM hallucination judger. These samples totally cover 28 domains, \emph{e.g.}, physics and religion.

\subsection{Human Data Cleaning}\label{sec:Hallucination Elimination}

We split the LLM-cleaned samples into the training, validation, and test sets with 84.6K, 1K and 1K samples, respectively.
In the validation and test sets, we balance the number of samples in each domain to ensure a comprehensive evaluation.
Furthermore, to mitigate the hallucination issue in the validation and test sets, we employ human annotators to manually cleanse their golden tables.

\begin{figure}[t!]
\centering
\includegraphics[scale=0.38]{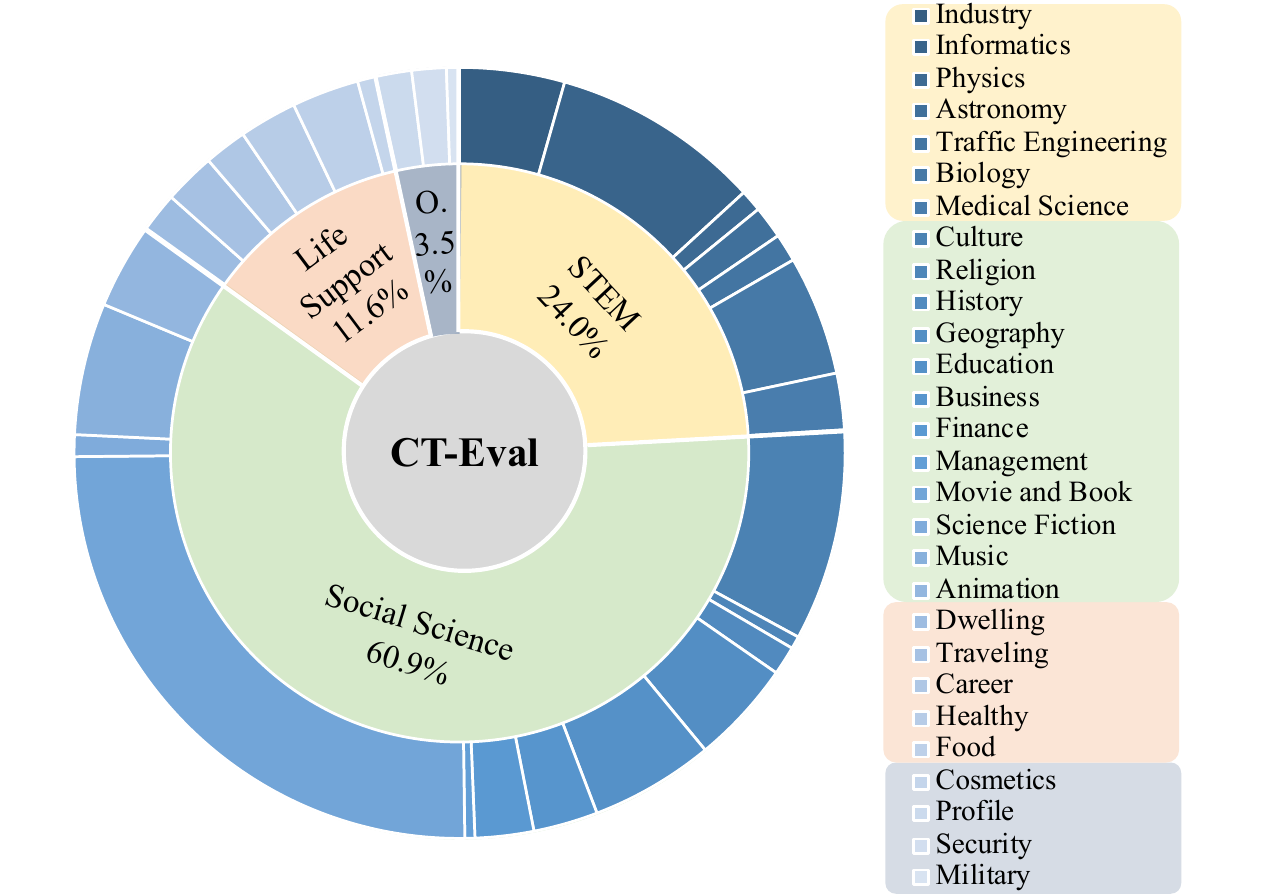}
\caption{Domain distribution in CT-EVAL}
\label{fig1}
\end{figure}

In this phase, we employ five human annotators and one data expert, all of whom are native Chinese speakers with advanced educational qualifications. The data expert is a researcher with extensive experience in IE research.
We first organize a tutorial for the five annotators to ensure alignment of annotation requirements. This includes clarifying the concept of the hallucination issue, emphasizing the additional information present in tables that cannot be directly extracted from the corresponding documents.
Next, for each document-table pair in the validation and test sets, three annotators are asked to remove the hallucination information from tables.
The data expert reviews 10\% of the manually cleaned samples from each annotator. If the accuracy rate falls below 95\%, the respective annotator will be required to redo the annotation.
Finally, for each sample, if the three manually cleaned results are consistent, the results are saved as the final data. Otherwise, the results are decided by a group meeting among all annotators and the data expert.

\subsection{Task Overview}\label{sec:Task definition}
Given a document $D = \{w_1, w_2, ..., w_{|D|}\}$, where $w_i$ is the $i$-th word in $D$, the text-to-table task aims to extract key information from $D$ and outputs a table $T = \{c_{0,0}, c_{0,1}, ..., c_{0,n}, c_{1,0}, c_{1,1}, ..., c_{1,n}, ..., \\ c_{i,j}, ..., c_{m, n}\}$, where $c_{0,k} (k \in \{0, 1, ..., n\})$ represents the column header and $c_{k,0} (k \in \{0, 1, ..., n\})$ denotes as the row header.
$c_{i,j} (i > 0, j > 0)$ represents the text of the cell in the $i$-th row and $j$-th column in the table with $m$ rows and $n$ columns.

\begin{table*}[t]
\centering
\resizebox{0.90\textwidth}{!}
{
\begin{tabular}{llrrlrr}
\toprule[1pt]
 \multicolumn{1}{c}{Dataset} & \multicolumn{1}{c}{Language} & \multicolumn{1}{c}{N. Entries} & \multicolumn{1}{c}{Avg. Length} & \multicolumn{1}{c}{N. Domains} & \multicolumn{1}{c}{Hallu. R} & \multicolumn{1}{c}{Avg. Cells} \\
\midrule[1pt]
E2E & English &42,061 & 90.58 & Restaurant & {4.17\%}& 4.46 \\ 
Rotowire & English &3,398 & 1311.01 & Sports & {6.17\%} & 40.49\\ 
WikiBio & English &582,659 & 416.71 & Biography & {8.67\%} & 4.19\\ 
WikiTableText & English &10,000 & 59.76 & Multiple Subclasses & {18.83\%}& 4.25\\
\midrule[1pt] 
CT-Eval(train) & Chinese &84,603 & 911.46 & 28 Subclasses & { 6.83\%}& 11.40 \\ 
CT-Eval(val.) & Chinese &1,000 & 813.32 & 28 Subclasses & { 1.00\%}& 10.78 \\ 
CT-Eval(test) & Chinese &1,000 & 845.22 & 28 Subclasses & { 1.50\%}& 10.86 \\ 
\bottomrule[1pt]
\end{tabular}
}
\caption{Data Statistics of CT-Eval and previous text-to-table datasets. ``\emph{N. Entries}'' denotes the number of entries. ``\emph{Hallu. R}'' indicates hallucination rate. ``\emph{Avg.Cells}'' indicates the average number of table cells }
\label{table1}
\end{table*}

\subsection{Data Statistics}
\label{sec:Dataset Statics}
We compare CT-Eval with the previous datasets across several metrics including language, number of entries, average length, number of domains, hallucination rate, and average number of cells. 

\begin{figure}[t!]
\centering
\includegraphics[scale=0.46]{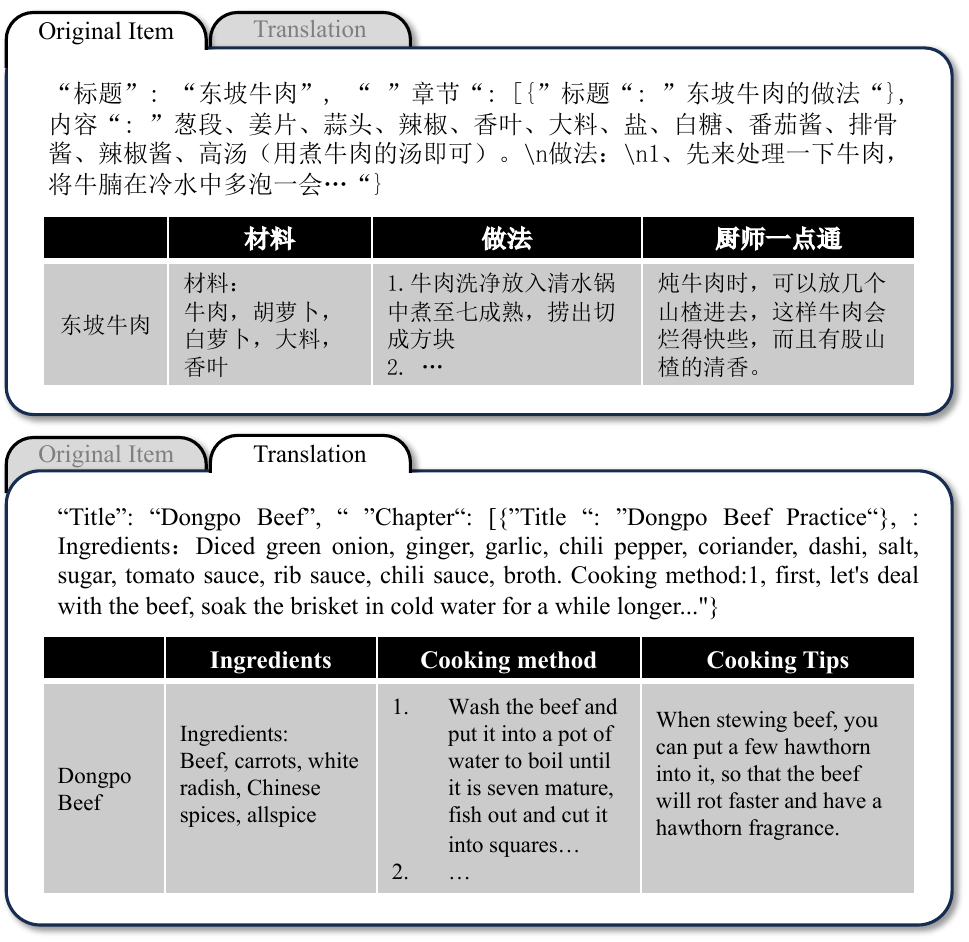}
\caption{Text-to-table example from CT-EVAL.}
\label{data_sampling}
\end{figure}

As shown in Table~\ref{table1}, compared to previous datasets that mostly focus on one domain, CT-Eval covers 28 domains, offering a valuable resource to evaluate the text-to-table capabilities of LLMs across multiple domains.
These 28 domains can be categorized into four branches: STEM, social science, life support and others.
To provide a comprehensive understanding of domain coverage in CT-Eval, we present the distributions of each domain in Figure~\ref{fig1}. Notably, the \emph{``Social Science''} branch exhibits the highest proportion, encompassing domains such as culture and religion.
Conversely, the \emph{``Other''} branch, which includes topics related to cosmetics and profiles, represents the smallest portion.
The \emph{``STEM''} (science, technology, engineering, and mathematics) branch features data related to topics such as physical sciences and astronomy.
The \emph{``Life Support''} branch is highly relevant to everyday life, comprises 11.60\% of the data, and primarily focuses on domains such as dwelling and health.
An illustration of a data sample from CT-Eval is depicted in Figure~\ref{data_sampling}.
Each document-table pair, akin to the example shown, consists of a textual description and a golden table with varying numbers of columns and rows.

To assess the quality of our dataset compared to previous ones, we randomly select 200 samples from the training, validation and test sets of CT-Eval, and previous E2E, Rotowire, WikiBio, and WikiTableText, respectively.
Then, we compute the hallucination rate for each dataset through human annotation.
Specifically, three annotators proficient in both English and Chinese judged whether the golden tables contained hallucination information following the guidance outlined in Section~\ref{sec:Hallucination Elimination}.
The hallucination rate for each dataset is determined by the average proportion of hallucinated samples judged by all three annotators.
We observe that the hallucination rate of the CT-Eval training set is 6.83\%, similar to Rotowire (6.17\%).
Moreover, with the help of our human cleaning, the hallucination rates in our test and validation sets decrease to 1.00\% and 1.50\%, respectively, significantly lower than those in other datasets.
Therefore, the reliability of our dataset can be verified.
Regarding domains, the E2E, Rotowire and WikiBio datasets focus on single domains, whereas WikiTableText and CT-Eval encompass multiple domains to ensure data diversity.
Concerning the length of the input documents, we note that previous E2E and WikiTableText primarily contain short documents with an average length of under 100 words, whereas documents in our dataset and Rotowire exceed  words or characters, presenting more complex inputs for text-to-table models.
In summary, CT-Eval stands as the sole dataset fulfilling criteria of data diversity, lengthy documents, and low hallucination.

\section{Experiment}

\subsection{Model Selection}
We conduct experiments using two types of backbones: closed-source and open-source LLMs.

\vspace{0.5ex}
\noindent \textbf{Closed-Source LLMs.}
(1) \texttt{GPT-3.5-turbo}~\cite{ChatGPT} is one of the LLMs from GPT-3.5 family developed by OpenAI. It is trained through reinforcement learning from human feedback (RLHF). Renowned for its robust ability as a general task solver, we use its newest version, \emph{i.e.}, \emph{gpt-3.5-turbo-1106}.
(2) \texttt{GPT-4}~\cite{achiam2023gpt} is the most advanced GPT version inherited from GPT-3.5 and achieves the state-of-the-art zero-shot performance across a wide range of NLP tasks. However, this model also suffers from hallucinations~\cite{achiam2023gpt}. We use its newest version, \emph{i.e.}, \emph{gpt-4-1106}.

\vspace{0.5ex}
\noindent \textbf{Open-Source LLMs.}
(3) \texttt{ChatGLM-6B}, (4) \texttt{ChatGLM2-6B} and (5) \texttt{ChatGLM3-6B}, are a series of bilingual dialogue LLMs based on the General Language Model (GLM) architecture~\cite{du2022glm}.
(6) \texttt{Baichuan2-7B-Chat} and (7) \texttt{Baichuan2-13B-Chat} is also trained with RLHF. This model shows its superior multi-lingual abilities in downstream tasks~\cite{yang2023baichuan}.
(8) \texttt{Llama-Chinese-2-7B} and (9) \texttt{LLama-Chinese-2-13B}~\cite{Chinese-LLaMA-Alpaca} are developed based on original Llama-2~\cite{touvron2023llama2} with an expansion of the Chinese vocabulary and incremental pre-training with large-scale Chinese corpus.
(10) \texttt{Qwen-7B-Chat} and (11) \texttt{Qwen-14B-Chat}~\cite{bai2023qwen} are two LLMs of Qwen family that have been pretrained with multilingual corpus with 2.4 trillion and 3 trillion, respectively, covering a wide range of domains.

\subsection{Experiment Setups}

\subsubsection{Text-to-table Prompt}

Recent studies~\cite{liu2023pre,gao2020making,shin2020autoprompt} demonstrate that an LLM can effectively leverage a small number of examples or task-specific instructions, often referred to as a ``prompt'', to guide its performance across a variety of tasks. This approach frequently leads to improved performance outcomes.

\begin{figure}[t]
\centering
\includegraphics[scale=0.5]{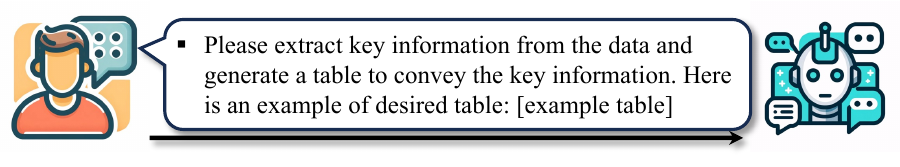}
\caption{Illustration of text-to-table prompt.}
\label{prompt}
\end{figure}

In our experiments, for both zero-shot and fine-tuning scenarios, we utilize a \emph{text-to-table prompt} designed to enhance model's capacity to construct a table from the provided text. The precise prompt we employ is illustrated in Figure~\ref{prompt}, where ``[example table]'' is an example of the golden table. This example table is retrieved from the training set of CT-Eval, providing LLMs with insight into the expected output styles.

\subsubsection{Implementation Details}

\noindent \textbf{Zero-Shot Prompting Details.}
(1) For the closed-source LLMs (\texttt{ChatGPT} and \texttt{GPT-4}), we access the official APIs provided by OpenAI\footnote{\url{https://openai.com/blog/openai-api}}, setting the \emph{temperature} to 0, while leaving the other hyper-parameters at their default settings.
(2) Regarding the open-source LLMs, we also set the temperature to 0, with the remaining hyper-parameters retained at their default values. Model inference is performed using a single NVIDIA Tesla V100 GPU with 32 GB of memory.

\vspace{0.5ex}
\noindent \textbf{Fine-Tuning Details} We set the number of training epochs as two and set the learning rate to 1e-5. Additionally, we implement a warm-up schedule~\cite{gotmare2018closer} during training. The optimizer employed is Adam~\cite{kingma2014adam}.
Furthermore, we utilize the DeepSpeed optimization~\cite{Rasley2020DeepSpeedSO}, and configure ZeRO-2 optimization for fine-tuning all LLMs.

\subsubsection{Metrics}

Following~\citet{wu2021text}, we evaluate the text-to-table performance based on the number of correct non-header cells.
Supposing a predicted table with $m$ rows and $n$ columns has $N_c=m\times n$ cells and the corresponding ground truth with $m^*$ rows and $n^*$ columns has $N^*_c=m^*\times n^*$ cells, we calculate the similarity $S$ between the predicted table and the ground truth as follows:

\begin{equation}
S = \frac{1}{N_c\times N^*_c} \sum_{i=1}^{m}{\sum_{j=1}^{n}{{\sum_{k=1}^{m^*}{{\sum_{l=1}^{n^*}{{\boldsymbol{O}(c_{i,j}, c^*_{k,l})}}}}}}}
\label{cell_sim}
\end{equation}

\noindent where $c_{i,j}$ and $c^*_{k,l}~(i,j,k,l > 0)$ refer to the predicted and ground-truth non-header cell content, respectively. $\boldsymbol{O}(\cdot)$ refers to there metrics used in our experiments:
(1) \textbf{E-Matching}: string-level exact matches are conducted between the texts of $c_{i,j}$ and $c^*_{k,l}$. 
(2) \textbf{Chrf-score}: the character-level $n$-gram ($n=6$) similarity between the texts of $c_{i,j}$ and $c^*_{k,l}$ are calculated.
(3) \textbf{BERT-score}: the vector similarity is operated between the BERT embeddings of $c_{i,j}$ and $c^*_{k,l}$.

\begin{table*}[ht]
\centering
\resizebox{1.00\textwidth}{!}
{
\begin{tabular}{l|l|ccc|ccc|ccc|ccc|ccc}
\toprule
\multicolumn{2}{c|}{\multirow{2}{*}{\textbf{Model}}} & \multicolumn{3}{c|}{\textbf{STEM}} &  \multicolumn{3}{c|}{\textbf{Social Sci.}} & \multicolumn{3}{c|}{\textbf{Life Supp.}} & \multicolumn{3}{c|}{\textbf{Others}} & \multicolumn{3}{c}{\textbf{Overall}} \\ 
\multicolumn{2}{c|}{} & P & R & F1 & P & R & F1 & P & R & F1 & P & R & F1 & P & R & F1 \\
\midrule
\multirow{11}{*}{\rotatebox{90}{E-matching}} & \multicolumn{1}{c|}{GPT3.5-turbo} & 6.08 & 17.18 & 8.08 & 0.60 & 7.14 & 1.10 & 0.00 & 0.00 & 0.00 & 0.00 & 0.00 & 0.00 & 2.09 & 5.52 & 2.87\\
& \multicolumn{1}{c|}{GPT-4} & \textbf{7.46} & \textbf{33.33} & \textbf{11.18} & \textbf{5.40} & \textbf{35.71} & \textbf{8.87} & \textbf{2.14} & \textbf{14.55} & \textbf{3.30} & 4.03 & \textbf{20.00} & 6.44 & \textbf{5.07} &    \textbf{22.10} & \textbf{7.50}\\ 
& \multicolumn{1}{c|}{ChatGLM-6B} & 0.00 & 0.00 & 0.00 & 0.00 & 0.00 & 0.00 & 0.00 & 0.00 & 0.00 & 0.00 & 0.00 & 0.00 & 0.00 & 0.00 & 0.00\\
& \multicolumn{1}{c|}{ChatGLM2-6B} & 0.00 & 0.00 & 0.00 & 0.00 & 0.00 & 0.00 & 0.00 & 0.00 & 0.00 & 0.00 & 0.00 & 0.00 & 0.00 & 0.00 & 0.00\\
& \multicolumn{1}{c|}{ChatGLM3-6B} & 0.00 & 0.00 & 0.00 & 0.00 & 0.00 & 0.00 & 0.00 & 0.00 & 0.00 & 0.00 & 0.00 & 0.00 & 0.00 & 0.00 & 0.00\\
 & \multicolumn{1}{c|}{Baichuan2-7B-Chat} & 1.39 & 2.78 & 1.85 & 0.00 & 0.00 & 0.00 & 0.02 & 0.61 & 0.05 & \textbf{6.67} & 13.33 & \textbf{8.89} & 2.01 & 5.24 & 2.74  \\
 & \multicolumn{1}{c|}{Baichuan2-13B-Chat} &1.95  &  4.52&   2.71& 0.00 &0.00  &0.00  & 0.00 & 0.00 & 0.00  &  0.00& 0.00 &0.00  & 0.48 & 1.19 & 0.57  \\
 & \multicolumn{1}{c|}{Llama-Chinese-2-7B} & 2.68 & 6.57 & 3.73 & 0.00 & 0.00 & 0.00 & 0.00 & 0.00 & 0.00 & 0.00 & 0.00 & 0.00 & 1.62 & 4.07 & 2.21\\
 & \multicolumn{1}{c|}{Llama-Chinese-2-13B} &  1.18 & 2.55 &  1.61 & 0.00 & 0.00 & 0.00 &   0.91&  0.91 &  0.91 & 0.00 &  0.00 & 0.00 &  1.13 &  2.70  &  1.47 \\
 & \multicolumn{1}{c|}{Qwen-7B-Chat} & 0.00 & 0.00 & 0.00 & 0.00 & 0.00 & 0.00 & 0.91  &  0.91 & 0.91 &0.00 & 0.00 & 0.00 & 0.57  & 1.03 &  0.69 \\
& \multicolumn{1}{c|}{Qwen-14B-Chat} & 1.00 &  2.08 & 1.35 & 0.00  &  0.00 & 0.00 & 0.00 & 0.00 & 0.00 & 0.00 & 0.00 & 0.00 & 0.23 & 0.62 &  0.29 \\
\hline
\multirow{11}{*}{\rotatebox{90}{Chrf-score}} & \multicolumn{1}{c|}{GPT3.5-turbo} & 23.54 & 59.00 & \textbf{31.58} & 10.22 & 63.19 & 15.90 & 9.81 & 30.28 & \textbf{12.56} & 13.89 & 51.90 & 20.88 & 15.96 & 48.60 & 21.71\\
& \multicolumn{1}{c|}{GPT-4} & 23.26 & \textbf{63.80} & 30.82 & 18.60 & \textbf{68.96} & 27.34 & 9.85 & \textbf{31.29} & 12.36 & 15.77 & \textbf{52.59} & 23.58 & \textbf{18.76} & \textbf{54.19} & \textbf{25.57}\\ 
& \multicolumn{1}{c|}{ChatGLM-6B} & 4.54 & 8.22 & 5.65 & 3.37 & 8.86 & 4.79 & 2.05 & 3.87 & 2.61 & 2.15 & 4.61 & 2.88 & 3.07 & 6.38 & 4.02\\
& \multicolumn{1}{c|}{ChatGLM2-6B} & 2.27 & 4.16 & 2.67 & 0.71 & 3.53 & 1.01 & 0.81 & 2.59 & 1.02 & 0.83 & 2.41 & 1.04 & 1.29 & 3.37 & 1.63  \\
& \multicolumn{1}{c|}{ChatGLM3-6B} & 2.22 & 6.45 & 2.96 & 1.23 & 4.63 & 1.79 & 1.42 & 4.81 & 1.96 & 1.21 & 6.02 & 1.95 & 1.55 & 5.25 & 2.16\\
 & \multicolumn{1}{c|}{Baichuan2-7B-Chat} & 13.18 & 22.57 & 15.31 & 4.74 & 17.80 & 6.79 & 7.43 & 11.83 & 7.59 & 13.39  & 42.24 & 18.82 & 11.5 & 25.55 & 14.35  \\
 & \multicolumn{1}{c|}{Baichuan2-14B-Chat} & 8.89 &  21.01 &  11.43 & 1.36 &  4.86 & 2.01 & 5.43 &  12.02 &  5.74 &  6.98 & 17.66 &  9.89 & 4.08 &   10.12&  4.83  \\
 & \multicolumn{1}{c|}{Llama-Chinese-2-7B} & \textbf{24.27} & 48.06 & 28.43 & \textbf{24.86} & 52.73 & \textbf{29.13} & 10.76 & 17.66 & 9.41 & 20.78 & 41.08 & 23.20 & 16.65 & 39.76 & 20.27 \\
 & \multicolumn{1}{c|}{Llama-Chinese-2-13B} &  16.79 & 34.15 & 20.60 &  7.09 & 34.74 & 10.89 &  5.81 & 13.21 & 7.04 & 5.43 & 12.02 &  5.74 & 12.18 & 33.59 & 15.81  \\
 & \multicolumn{1}{c|}{Qwen-7B-Chat} & 17.18 &  23.09 & 17.01 & 17.18 & 23.09 & 17.01 & \textbf{14.66} & 11.93 & 11.92 &  19.28 & 39.35 & 24.05 & 17.85 & 25.08 &  18.14 \\
& \multicolumn{1}{c|}{Qwen-14B-Chat} & 10.51 &  15.71 & 10.09 &  12.82 & 28.71 & 14.65 & 11.65 &  9.70 &  8.96 & \textbf{25.01} &  32.39 & \textbf{26.52} & 16.62 &  24.25 & 16.97 \\
\hline
\multirow{11}{*}{\rotatebox{90}{BERT-score}} & \multicolumn{1}{c|}{GPT3.5-turbo} & 25.64 & 63.09 & 34.35 & 14.98 &  64.36 & 22.78 & 11.54 & 33.61 &  15.34 & 14.22 & 51.95 & 21.37 & 16.82 & 49.54 & 23.06\\
& \multicolumn{1}{c|}{GPT-4} & \textbf{28.54} & \textbf{72.81} & \textbf{37.56} & 21.83 & \textbf{72.91} & 32.46 & 11.93 & \textbf{36.31} & \textbf{15.84} & 17.16 & \textbf{64.85} & 26.36 & \textbf{20.83} & \textbf{61.02} & \textbf{28.85} \\
& \multicolumn{1}{c|}{ChatGLM-6B} & 3.26 & 4.98 & 3.61 & 2.19 & 6.40 & 3.21 & 3.97 & 2.02 & 2.52 & 1.15 & 3.40 & 1.67 & 2.52 & 5.57 & 3.24\\
& \multicolumn{1}{c|}{ChatGLM2-6B} & 2.15 & 5.07 & 2.44 & 1.12 & 9.05 & 1.94 & 1.89 & 6.24 & 2.52 & 2.22 & 5.24 & 2.70 & 2.07 & 7.29 & 2.84\\
& \multicolumn{1}{c|}{ChatGLM3-6B} & 6.49 & 20.29 & 8.73 & 2.57 & 14.51 & 4.18 & 3.07 & 14.27 & 4.63 & 3.99 & 19.34 & 6.26 & 4.11 & 17.05 & 6.03\\
 & \multicolumn{1}{c|}{Baichuan2-7B-Chat} & 13.26 & 24.73 & 16.60 & 6.81 & 21.92 & 9.93 & 10.29  & 17.21 & 11.08 & 15.47 & 47.12 & 21.95 & 12.64 & 28.98 & 16.23\\
 & \multicolumn{1}{c|}{Baichuan2-14B-Chat} & 10.92 &  26.68 & 14.81 & 2.81&  14.10  & 4.58 &   5.38&13.87  &6.10  & 7.53 & 19.05 & 10.67 &  12.32& 27.67 & 15.56   \\
 & \multicolumn{1}{c|}{Llama-Chinese-2-7B} & 26.46 & 48.85 & 31.26 & \textbf{29.53} & 55.04 & \textbf{35.29} & 13.65 & 21.10 & 12.74 & 21.98 & 45.49 &  24.93 & 18.52 & 41.08 & 22.53  \\
 & \multicolumn{1}{c|}{Llama-Chinese-2-13B} &18.88  & 38.95 & 22.38 & 10.04  &  37.90& 14.84  & 8.93 & 19.30 & 11.03 & 6.85 & 23.43 &  9.82 &15.11  & 38.17 &  19.39 \\
 & \multicolumn{1}{c|}{Qwen-7B-Chat} &  20.08 & 28.20 & 20.87 & 24.81 & 36.03 & 26.20 & \textbf{16.78} & 15.70 &  14.78 & 23.64 & 49.01 &  29.85 & 19.39 & 27.28 &  19.94 \\
& \multicolumn{1}{c|}{Qwen-14B-Chat} &11.79 &  12.93 & 10.72 &  15.30 & 32.81 & 18.23 & 12.24 &  11.68 & 10.10 & \textbf{28.10} &  39.50 & \textbf{30.25} &18.60 & 26.78 & 19.11 \\
\hline
\end{tabular}
}
\caption{Benchmark LLMs on CT-Eval in the zero-shot manner, where the best scores are in \textbf{bold}. \emph{STEM} denotes Science, Technology, Engineering and Mathematics subject; \emph{Social Sci.} denotes Social Science; \emph{Life Supp.} denotes Life Support. P, R and F1 indicates precision, recall and F1-score, respectively.}
\label{zero-shot}
\end{table*}

\begin{table*}[t!]
\centering

\resizebox{0.80\textwidth}{!}
{
\begin{tabular}{l|r|r|r|r|r|r}
\hline
 LLMs  & Error Rate  & \textsc{Error A} & \textsc{Error B} &  \textsc{Error C}  &  \textsc{Error D} & \textsc{Error E}\\
\hline
GPT-3.5-turbo & 74\% &28\% &44\% & 2\% & 0\% & 0\% \\
GPT-4 & 58\% &4\% & 34\% & 6\% & 14\% &0\% \\
Baichuan2-7B-Chat & 94\% & 70\% & 20\% & 2\% & 2\% & 0\% \\
Llama-Chinese-2-7B & 88\% &0\% & 73\% & 3\% &7\%& 5\%\\
Qwen-7B-Chat & 96\%& 72\% &16\% & 6\% & 2\% & 0\% \\
\hline
\end{tabular}
}
\caption{Bad case analysis for zero-shot. \textsc{Error A}: containing some hallucinations; \textsc{Error B}: the outputs are over similar to the inputs; \textsc{Error C}: returning completely irrelevant information; \textsc{Error D}: the summarized information is inconsistent with the golden table; \textsc{Error E}: returning a proper table with duplication.}
\label{table6}
\end{table*}

\begin{table*}[ht]
\centering
\resizebox{1.00\textwidth}{!}
{
\begin{tabular}{l|l|rrr|rrr|rrr|rrr|rrr}
\hline
\multicolumn{2}{c|}{\multirow{2}{*}{\textbf{Model}}} & \multicolumn{3}{c|}{\textbf{STEM}} &  \multicolumn{3}{c|}{\textbf{Social Sci.}} & \multicolumn{3}{c|}{\textbf{Life Supp.}} & \multicolumn{3}{c|}{\textbf{Others}} & \multicolumn{3}{c}{\textbf{Overall}} \\
\multicolumn{2}{c|}{} & P & R & F1 & P & R & F1 & P & R & F1 & P & R & F1 & P & R & F1 \\
\hline
\multicolumn{1}{c|}{\multirow{10}{*}{\rotatebox{90}{E-matching}}}
& \multicolumn{1}{c|}{ChatGLM-6B} & \textbf{30.57} & 25.55 & \textbf{27.25} & \textbf{1.79} & \textbf{7.14} & \textbf{2.86} & 8.78 & 24.30 & 12.31 & 0.00 & 0.00 & 0.00 & \textbf{15.13} & \textbf{22.84} & \textbf{15.96}\\

 & \multicolumn{1}{c|}{ChatGLM2-6B} & 24.42 & 21.42 &  22.60 & 0.00 & 0.00 & 0.00 &  \textbf{9.57}  &  30.12 & \textbf{13.96} & \textbf{12.30} & \textbf{48.85} & \textbf{19.06} &  10.83 & 16.25 & 11.72 \\
 & \multicolumn{1}{c|}{ChatGLM3-6B} & 25.01 &  22.23 &  23.33 & 0.00 & 0.00 & 0.00 &  7.56 & 24.42 & 11.02 & 1.11 & 3.33 &  1.67 & 10.92 & 15.97 & 11.67 \\
 & \multicolumn{1}{c|}{Baichuan2-7B-Chat} & 19.48 & 10.99 & 13.85 & 0.00 & 0.00 & 0.00 & 3.03 & 5.15 & 3.58 & 0.00 & 0.00 & 0.00 & 12.14 & 13.15 & 11.26\\
 & \multicolumn{1}{c|}{Baichuan2-13B-Chat} & 26.34 & 23.30 & 24.38 & 0.00 & 0.00 & 0.00 & 8.89 &  27.52 & 12.85 & 0.00 & 0.00 & 0.00 & 10.60 &  15.44 & 11.32 \\
 & \multicolumn{1}{c|}{Llama-Chinese-2-7B} & 22.52 & 15.61 & 18.16 & 0.00 & 0.00 & 0.00 & 8.18 & 22.85 & 11.41 & 0.00 & 0.00 & 0.00 & 13.30 & 17.68 & 13.26\\
 & \multicolumn{1}{c|}{Llama-Chinese-2-13B} &  24.87 &  23.23 & 22.75 & 0.00 & 0.00 & 0.00 &8.96 & \textbf{31.09} & 13.41 & 0.00 &  0.00 & 0.00 &  10.59 &  15.60 &  11.30 \\
 & \multicolumn{1}{c|}{Qwen-7B-Chat} & 27.47 &  \textbf{28.46} & 26.27 & 0.00 & 0.00 & 0.00 &  9.15 & 30.18 & 13.46 & 1.67 & 3.33 &  2.22 & 10.99 & 16.42 &  11.77\\
& \multicolumn{1}{c|}{Qwen-14B-Chat} & 25.61 & 25.18 & 24.00 & 0.00 & 0.00 & 0.00 &  8.03  & 26.85 & 11.91 & 0.00 & 0.00 & 0.00 &  9.95 & 14.73 & 10.58\\
\hline
\multicolumn{1}{c|}{\multirow{10}{*}{\rotatebox{90}{Chrf-score}}} 
& 
 \multicolumn{1}{c|}{ChatGLM-6B} & 41.77 & 40.77 & 39.74 & 11.24 & 36.19 & 16.68 & 21.11 & 47.76 & 27.27 & 7.23 & 22.71 & 10.54 & 25.03 & 42.10 & 28.19 \\
 & \multicolumn{1}{c|}{ChatGLM2-6B} & 39.79 &  55.73 & 42.41 & 15.05 & 64.70 &  23.94 & \textbf{25.12} & 65.45 &  34.74 & 12.30 &  48.85 & 19.06 & 26.33 &  60.24 & 33.41 \\
 & \multicolumn{1}{c|}{ChatGLM3-6B} & 40.73 &  58.31 &  43.62 &  15.87 & 62.06 & 24.39 & 22.47 &  61.67 & 31.84 & 17.15 & 51.11 &  24.60 & 26.47 & 59.34 &  33.43 \\
 & \multicolumn{1}{c|}{Baichuan2-7B-Chat} & 34.79 & 24.50 & 27.55 & 7.31 & 17.27 & 10.06 & 16.35 & 29.64 & 19.95 & 7.57 & 15.91 & 10.09 & 23.73 & 34.13 & 25.08 \\
& \multicolumn{1}{c|}{Baichuan2-13B-Chat} & 38.70 & 53.98 & 41.34 & 16.42 &  66.53 & 25.95 &  8.89 &  27.52 & 12.85 & 13.93 &  46.10 & 20.06 &  26.49 & 59.39 & 33.49 \\
 & \multicolumn{1}{c|}{Llama-Chinese-2-7B} & 36.41 & 31.68 & 31.97 & 6.08 & 21.53 & 9.37 & 19.69 & 46.15 & 25.90 & 5.96 & 15.77 & 8.44 & 22.49 & 35.59 & 24.50 \\
 & \multicolumn{1}{c|}{Llama-Chinese-2-13B} &   40.45 & \textbf{60.11} & 43.63 & \textbf{17.94} & \textbf{67.86} & \textbf{27.94} & 24.83 & \textbf{66.25} & \textbf{34.94} & 13.89 & 45.73 & 20.55 & \textbf{26.61} & 59.85 & 33.64 \\
 & \multicolumn{1}{c|}{Qwen-7B-Chat} & \textbf{42.29} & 59.77 & \textbf{45.62} &  16.85 & 63.38 & 26.30 & 24.91 &  65.26 &  34.62 & \textbf{17.35} & \textbf{57.06} & \textbf{25.78} & 26.58 & \textbf{60.33} & \textbf{33.67} \\
 & \multicolumn{1}{c|}{Qwen-14B-Chat} & 40.57 & 57.64 & 43.30 &  15.70 &  63.35 & 24.85 & 23.55 &63.64 & 32.99  & 13.61 & 47.50 &  20.10 & 25.51 & 59.05 & 32.46 \\
\hline
\multicolumn{1}{c|}{\multirow{10}{*}{\rotatebox{90}{Finetune}}} 
& \multicolumn{1}{c|}{ChatGLM-6B} & \textbf{43.60} & 44.18 & \textbf{42.39} & 11.32 & 39.13 & 18.13 & 24.20 & 51.11 & 30.36 & 9.30 & 25.58 & 12.80 & \textbf{27.03} & 45.36 & \textbf{31.28} \\
 & \multicolumn{1}{c|}{ChatGLM2-6B} & 37.74 &  49.29 & 40.40 & 12.86 & 51.74 & 20.04 & \textbf{24.71} & 
 59.39 & 33.14 & 10.58 & 36.03 & 15.44 & 24.63 & 50.99 &  30.57 \\
 & \multicolumn{1}{c|}{ChatGLM3-6B} & 37.85 &  50.44 &  40.71 & 12.49 & 49.48 & 19.51 &  22.83 &  56.74 & 30.98 & 13.52 & 37.51 &  18.94 & 24.75 & 50.47 & 30.58 \\
 & \multicolumn{1}{c|}{Baichuan2-7B} &  37.62 & 30.52 & 32.34 & 9.72 & 23.24 & 13.39 & 17.70 & 30.28 & 20.91 & 10.83 & 24.17 & 14.73 & 24.96 & 29.49 & 26.74 \\
 & \multicolumn{1}{c|}{Baichuan2-13B} &  35.53 &  46.58 & 37.96 & 14.28 & \textbf{56.03} & 22.39 & 23.38 & 56.79 &  31.45 & 10.01 & 29.69 & 14.08 & 10.60 & 15.44 &  11.32 \\
 & \multicolumn{1}{c|}{Llama-Chinese-2-7B} & 39.42 & 25.89 & 36.58 & 7.22 & 25.89 & 11.16 & 21.80 & 47.28 & 27.74 & 9.37 & 18.53 & 11.77 & 24.98 & 39.29 & 27.66 \\
 & \multicolumn{1}{c|}{Llama-Chinese-2-13B} &   38.00 &  51.45 &  40.93 & \textbf{15.27}  & 54.60 & \textbf{23.38} & 24.43 & \textbf{60.85} & \textbf{33.37} & 14.47 &  33.09 & 19.03 & 24.90 &   50.89 &  30.78 \\
 & \multicolumn{1}{c|}{Qwen-7B-Chat} &  38.93 & \textbf{52.41} &  41.95 & 13.52 & 50.11 & 20.97 & 24.27 &  59.05 &  32.67 & \textbf{16.92} & \textbf{45.65} & \textbf{23.60} & 24.93 &  \textbf{51.40} &  30.93 \\
& \multicolumn{1}{c|}{Qwen-14B-Chat} & 37.04 & 49.78 &  39.27 & 12.62 & 49.15 &  19.68 &  23.49 &  56.37 & 31.70 &  11.27 & 34.84 & 15.97 & 23.98 &  49.92 &  29.70 \\
\hline
\end{tabular}
}
\caption{Enhance the performance of the LLMs through fine-tuning, where the best scores are in \textbf{bold}.}
\label{finetune}
\end{table*}

\begin{table*}[t!]
\centering

\resizebox{0.75\textwidth}{!}
{
\begin{tabular}{l|r|r|r|r|r|r}
\hline
 LLMs  & Error Rate & \textsc{Error A} & \textsc{Error B} & \textsc{Error C}  &  \textsc{Error D}  & \textsc{Error E} \\   
\hline
ChatGLM-6B & 18\%& 2\% & 0\% &4\% & 12\% &  0\%  \\
Baichuan2-7B & 30\% & 14\% & 0\% & 4\% & 12\% &0\%  \\
Llama-Chinese-2-7B & 22\% &12\% & 0\% & 0\% & 10\% & 0\%  \\
\hline
\end{tabular}
}

\caption{Bad case analysis of fine-tuning.}
\label{table7}
\end{table*}

\section{Results \& Analyses}

\subsection{Zero-shot Results}
Table~\ref{zero-shot} lists the zero-shot performance of closed-source and open-source LLMs on CT-Eval.
In addition to F1-score, Table~\ref{zero-shot} also provides insights into precisions and recalls. In the text-to-table task, high recalls suggest that the LLM's ability to derive most of the target information present in the golden table; in contrast, low precisions may stem from the LLM over-collecting or creating irrelevant information which can be regarded as hallucination.
We analyze the results from the following aspects:

\paragraph{GPT-3.5-turbo vs. GPT-4}
It can be observed that both GPT-4 and GPT-3.5-turbo outperform the other open-source LLMs in most metrics. Compared with GPT-3.5-turbo, GPT-4 always shows a superior performance, notably achieving a 4.63-point improvement in the overall F1-score. The \emph{Social Science} category has the highest improvement of 7.77 points in term of the F1-score brought by GPT-4. These results correspond with OpenAI's reported performance comparisons between the two LLMs~\cite{achiam2023gpt}.

\paragraph{Open-source LLMs}
As summarized in Table~\ref{zero-shot}, among the open-scoure LLMs, Baichuan2-7B-Chat outperforms the other LLMs in terms of overall F1-score using E-matching, while Llama-Chinese-2-7B demonstrates the best performance in terms of overall F1-score using Chrf- and BERT-score. These LLMs vary in performance on different categories, with Llama-Chinese-2-7B outperforming the others in most cases. Notably, in the categories of \emph{STEM} and \emph{Social Science}, Llama-Chinese-2-7B exhibits 10.39-point and 9.09-point improvements over the second-best open-source LLM in terms of BERT-score, respectively.

Compared with GPT-4, Baichuan2-7B-Chat experiences a significant performance gap of 4.76 points using E-matching, and Llama-Chinese-2-7B experiences gaps of 5.30 and 6.32 points using Chrf- and BERT-score metrics, respectively. In the categories of \emph{Social Science} and \emph{Others}, when using Chrf- and BERT-score metrics, Llama-Chinese-2-7B and Qwen-14B-Chat even outperform GPT-4 in the F1-score. This suggests that these LLMs may benefit from the pretraining through the corresponding supplemented corpus in Chinese.

\paragraph{Bad Case Analysis}
To investigate the causes behind the aforementioned results, we additionally conduct the bad case analysis. Specifically, we select 100 samples of tables generated by zero-shot and manually perform the bad case selections. We classify a sample as erroneous if it falls into any of the types labeled from \textsc{Error A} to \textsc{Error E}: \textsc{Error A} signifies that containing some hallucinations; \textsc{Error B} occurs when the outputs of the LLM are overly similar to the inputs; \textsc{Error C} denotes that the LLM returns completely irrelevant information; \textsc{Error D} indicates that the LLM summarizes inconsistent information with the golden table; \textsc{Error E} arises when the LLM generates a proper table however with duplicated information. The total error rate and error rate of each type are shown in Table~\ref{table6}. For closed-source LLMs, GPT-4 has a lower total error rate than GPT-3.5-turbo, and GPT-4 suppresses more hallucinations. Both the LLMs suffer from \textsc{Error B}, suggesting a lack of effective understanding of the objective of the text-to-table tasks. This observation could also elucidate the high recall rates.

As for the open-souce LLMs, it can be seen that the error rate of Llama-Chinese-2-7B is the lowest among the open-source LLMs, however, it is still 14\% higher than that of GPT-3.5-turbo. Similar to the OpenAI LLMs, \textsc{Error B} occurs most frequently in Llama-Chinese-2-7B. Conversely, Baichuan2-7B-Chat and Qwen-7B-Chat mainly suffer from the hallucinations.

\subsection{Fine-tuning Results}
We fine-tune open-source LLMs using the CT-Eval training set. Table~\ref{table7} shows the evaluation results.

Compared with the zero-shot LLMs, there is a significant improvement in the performance of the open-source LLMs. ChatGLM-6B demonstrates the highest overall F1-scores for E-matching and BERT-score metrics, exhibiting increments of 15.96 and 28.04 points, respectively; Qwen-7B-Chat achieves the highest overall F1-score for Chrf-score metric with an increment of 15.53 points. Notably, after fine-tuning, ChatGLM-6B outperforms GPT-4 by a substantial margin across all the similarity metrics. These findings suggest that the open-source LLMs may not naturally excel in supporting complex IE tasks, and that the CT-Eval training set can effectively enhance the text-to-table ability of LLMs. This could also guide the future LLM research that the specific complex IE tasks should be considered in their developments.

\paragraph{Bad Case Analysis}
Furthermore, we investigate the possible changes in the types of the bad cases after fine-tuning, as illustrated in Table~\ref{table7}. The \textsc{Error B} and \textsc{Error E} types are no longer exists. The error rates of all the LLMs decreases, with ChatGLM-6B exhibiting the lowest error rate. While Baichuan2-7B and Llama-Chinese-2-7B primarily experience hallucinations. In contrast, ChatGLM-6B successfully suppressed hallucinations; however, its ability to precisely extract and organize information may require further enhancement through additional paradigms.

\section{Conclusion}
To assess the LLMs in the text-to-table tasks and further enhance the corresponding performance, we propose the first Chinese text-to-table dataset, CT-Eval, encompassing 28 domains. To minimize hallucination, LLM data cleaning and human cleaning are employed. Our comprehensive experiments cover closed- and open-source LLMs in both zero-shot and fine-tuning settings.
Experimental results demonstrate that in zero-shot settings, GPT-4 outperforms best among all LLMs. After fine tuning the open-source LLMs, significant performance improvements are observed, surpassing GPT-4 by considerable margins.
Bad case analysis further reveals the persistence of hallucinations in both zero-shot and fine-tuning settings. These findings indicate that current LLMs face challenges in text-to-table tasks, positioning CT-Eval as an effective benchmark for evaluating text-to-table tasks and valuable resource enhancing LLM performance. 

\section*{Limitations}
While we propose the first Chinese text-to-table benchmark dataset, there are some limitations worth considering in future work:
(1) CT-Eval only adopts human cleaning in the validation and test sets. Thus, the training set still involves hallucination issue.
(2) The input documents in CT-Eval only involve textual information while ignore the multi-modal information like figures. Future work could explore the multi-modal text-to-table performance beyond CT-Eval.

\section*{Ethical Considerations}

In this paper, we propose the CT-Eval dataset.
During human cleaning process, the salary for each human annotator is determined by the average time of annotation and local labor compensation standards.
For Baidu Baike, we utilize the data dumps provided by ~\citet{xu2017cn}. According to their official website's description, this dumps is authorized for academic usage.\footnote{\url{http://kw.fudan.edu.cn/cndbpedia/download/}}

\bibliography{custom}

\begin{thebibliography}{33}
\expandafter\ifx\csname natexlab\endcsname\relax\def\natexlab#1{#1}\fi

\bibitem[{Bai et~al.(2023{\natexlab{a}})Bai, Bai, Chu, Cui, Dang, Deng, Fan, Ge, Han, Huang et~al.}]{bai2023qwen}
Jinze Bai, Shuai Bai, Yunfei Chu, Zeyu Cui, Kai Dang, Xiaodong Deng, Yang Fan, Wenbin Ge, Yu~Han, Fei Huang, et~al. 2023{\natexlab{a}}.
\newblock Qwen technical report.
\newblock \emph{arXiv preprint arXiv:2309.16609}.

\bibitem[{Bai et~al.(2023{\natexlab{b}})Bai, Lv, Zhang, Lyu, Tang, Huang, Du, Liu, Zeng, Hou et~al.}]{bai2023longbench}
Yushi Bai, Xin Lv, Jiajie Zhang, Hongchang Lyu, Jiankai Tang, Zhidian Huang, Zhengxiao Du, Xiao Liu, Aohan Zeng, Lei Hou, et~al. 2023{\natexlab{b}}.
\newblock Longbench: A bilingual, multitask benchmark for long context understanding.
\newblock \emph{arXiv preprint arXiv:2308.14508}.

\bibitem[{Bao et~al.(2018)Bao, Tang, Duan, Yan, Lv, Zhou, and Zhao}]{bao2018table}
Junwei Bao, Duyu Tang, Nan Duan, Zhao Yan, Yuanhua Lv, Ming Zhou, and Tiejun Zhao. 2018.
\newblock Table-to-text: Describing table region with natural language.
\newblock In \emph{Proceedings of the AAAI conference on artificial intelligence}, volume~32.

\bibitem[{Cui et~al.(2023)Cui, Yang, and Yao}]{Chinese-LLaMA-Alpaca}
Yiming Cui, Ziqing Yang, and Xin Yao. 2023.
\newblock \href {https://arxiv.org/abs/2304.08177} {Efficient and effective text encoding for chinese llama and alpaca}.
\newblock \emph{arXiv preprint arXiv:2304.08177}.

\bibitem[{Du et~al.(2022)Du, Qian, Liu, Ding, Qiu, Yang, and Tang}]{du2022glm}
Zhengxiao Du, Yujie Qian, Xiao Liu, Ming Ding, Jiezhong Qiu, Zhilin Yang, and Jie Tang. 2022.
\newblock Glm: General language model pretraining with autoregressive blank infilling.
\newblock In \emph{Proceedings of the 60th Annual Meeting of the Association for Computational Linguistics (Volume 1: Long Papers)}, pages 320--335.

\bibitem[{Gao et~al.(2023)Gao, Zhao, Yu, and Xu}]{gao2023exploring}
Jun Gao, Huan Zhao, Changlong Yu, and Ruifeng Xu. 2023.
\newblock Exploring the feasibility of chatgpt for event extraction.
\newblock \emph{arXiv preprint arXiv:2303.03836}.

\bibitem[{Gao et~al.(2020)Gao, Fisch, and Chen}]{gao2020making}
Tianyu Gao, Adam Fisch, and Danqi Chen. 2020.
\newblock Making pre-trained language models better few-shot learners.
\newblock \emph{arXiv preprint arXiv:2012.15723}.

\bibitem[{Gonz{\'a}lez-Gallardo et~al.(2023)Gonz{\'a}lez-Gallardo, Boros, Girdhar, Hamdi, Moreno, and Doucet}]{gonzalez2023yes}
Carlos-Emiliano Gonz{\'a}lez-Gallardo, Emanuela Boros, Nancy Girdhar, Ahmed Hamdi, Jose~G Moreno, and Antoine Doucet. 2023.
\newblock Yes but.. can chatgpt identify entities in historical documents?
\newblock \emph{arXiv preprint arXiv:2303.17322}.

\bibitem[{Gotmare et~al.(2018)Gotmare, Keskar, Xiong, and Socher}]{gotmare2018closer}
Akhilesh Gotmare, Nitish~Shirish Keskar, Caiming Xiong, and Richard Socher. 2018.
\newblock A closer look at deep learning heuristics: Learning rate restarts, warmup and distillation.
\newblock \emph{arXiv preprint arXiv:1810.13243}.

\bibitem[{Kingma and Ba(2014)}]{kingma2014adam}
Diederik~P Kingma and Jimmy Ba. 2014.
\newblock Adam: A method for stochastic optimization.
\newblock \emph{arXiv preprint arXiv:1412.6980}.

\bibitem[{Lebret et~al.(2016)Lebret, Grangier, and Auli}]{lebret2016neural}
R{\'e}mi Lebret, David Grangier, and Michael Auli. 2016.
\newblock Neural text generation from structured data with application to the biography domain.
\newblock \emph{arXiv preprint arXiv:1603.07771}.

\bibitem[{Lewis et~al.(2019)Lewis, Liu, Goyal, Ghazvininejad, Mohamed, Levy, Stoyanov, and Zettlemoyer}]{lewis2019bart}
Mike Lewis, Yinhan Liu, Naman Goyal, Marjan Ghazvininejad, Abdelrahman Mohamed, Omer Levy, Ves Stoyanov, and Luke Zettlemoyer. 2019.
\newblock Bart: Denoising sequence-to-sequence pre-training for natural language generation, translation, and comprehension.
\newblock \emph{arXiv preprint arXiv:1910.13461}.

\bibitem[{Li et~al.(2023)Li, Wang, Shao, Zheng, Wang, and Su}]{li2023sequence}
Tong Li, Zhihao Wang, Liangying Shao, Xuling Zheng, Xiaoli Wang, and Jinsong Su. 2023.
\newblock A sequence-to-sequence\&set model for text-to-table generation.
\newblock \emph{arXiv preprint arXiv:2306.00137}.

\bibitem[{Liu et~al.(2023)Liu, Yuan, Fu, Jiang, Hayashi, and Neubig}]{liu2023pre}
Pengfei Liu, Weizhe Yuan, Jinlan Fu, Zhengbao Jiang, Hiroaki Hayashi, and Graham Neubig. 2023.
\newblock Pre-train, prompt, and predict: A systematic survey of prompting methods in natural language processing.
\newblock \emph{ACM Computing Surveys}, 55(9):1--35.

\bibitem[{Novikova et~al.(2017)Novikova, Du{\v{s}}ek, and Rieser}]{novikova2017e2e}
Jekaterina Novikova, Ond{\v{r}}ej Du{\v{s}}ek, and Verena Rieser. 2017.
\newblock The e2e dataset: New challenges for end-to-end generation.
\newblock \emph{arXiv preprint arXiv:1706.09254}.

\bibitem[{OpenAI(2022)}]{ChatGPT}
OpenAI. 2022.
\newblock Introducing chatgpt.
\newblock \url{https://openai.com/blog/chatgpt}.

\bibitem[{OpenAI(2023)}]{achiam2023gpt}
OpenAI. 2023.
\newblock Gpt-4 technical report.
\newblock \emph{ArXiv}, abs/2303.08774.

\bibitem[{Pietruszka et~al.(2022)Pietruszka, Turski, Borchmann, Dwojak, Pa{\l}ka, Szyndler, Jurkiewicz, and Garncarek}]{pietruszka2022stable}
Micha{\l} Pietruszka, Micha{\l} Turski, {\L}ukasz Borchmann, Tomasz Dwojak, Gabriela Pa{\l}ka, Karolina Szyndler, Dawid Jurkiewicz, and {\L}ukasz Garncarek. 2022.
\newblock Stable: Table generation framework for encoder-decoder models.
\newblock \emph{arXiv preprint arXiv:2206.04045}.

\bibitem[{Powalski et~al.(2021)Powalski, Borchmann, Jurkiewicz, Dwojak, Pietruszka, and Pa{\l}ka}]{powalski2021going}
Rafa{\l} Powalski, {\L}ukasz Borchmann, Dawid Jurkiewicz, Tomasz Dwojak, Micha{\l} Pietruszka, and Gabriela Pa{\l}ka. 2021.
\newblock Going full-tilt boogie on document understanding with text-image-layout transformer.
\newblock In \emph{Document Analysis and Recognition--ICDAR 2021: 16th International Conference, Lausanne, Switzerland, September 5--10, 2021, Proceedings, Part II 16}, pages 732--747. Springer.

\bibitem[{Raffel et~al.(2020)Raffel, Shazeer, Roberts, Lee, Narang, Matena, Zhou, Li, and Liu}]{raffel2020exploring}
Colin Raffel, Noam Shazeer, Adam Roberts, Katherine Lee, Sharan Narang, Michael Matena, Yanqi Zhou, Wei Li, and Peter~J Liu. 2020.
\newblock Exploring the limits of transfer learning with a unified text-to-text transformer.
\newblock \emph{The Journal of Machine Learning Research}, 21(1):5485--5551.

\bibitem[{Rasley et~al.(2020)Rasley, Rajbhandari, Ruwase, and He}]{Rasley2020DeepSpeedSO}
Jeff Rasley, Samyam Rajbhandari, Olatunji Ruwase, and Yuxiong He. 2020.
\newblock \href {https://api.semanticscholar.org/CorpusID:221191193} {Deepspeed: System optimizations enable training deep learning models with over 100 billion parameters}.
\newblock \emph{Proceedings of the 26th ACM SIGKDD International Conference on Knowledge Discovery \& Data Mining}.

\bibitem[{Shin et~al.(2020)Shin, Razeghi, Logan~IV, Wallace, and Singh}]{shin2020autoprompt}
Taylor Shin, Yasaman Razeghi, Robert~L Logan~IV, Eric Wallace, and Sameer Singh. 2020.
\newblock Autoprompt: Eliciting knowledge from language models with automatically generated prompts.
\newblock \emph{arXiv preprint arXiv:2010.15980}.

\bibitem[{Touvron et~al.(2023)Touvron, Martin, Stone, Albert, Almahairi, Babaei, Bashlykov, Batra, Bhargava, Bhosale et~al.}]{touvron2023llama2}
Hugo Touvron, Louis Martin, Kevin Stone, Peter Albert, Amjad Almahairi, Yasmine Babaei, Nikolay Bashlykov, Soumya Batra, Prajjwal Bhargava, Shruti Bhosale, et~al. 2023.
\newblock Llama 2: Open foundation and fine-tuned chat models, 2023.
\newblock \emph{URL https://arxiv. org/abs/2307.09288}.

\bibitem[{Wang et~al.(2018)Wang, Singh, Michael, Hill, Levy, and Bowman}]{wang2018glue}
Alex Wang, Amanpreet Singh, Julian Michael, Felix Hill, Omer Levy, and Samuel~R Bowman. 2018.
\newblock Glue: A multi-task benchmark and analysis platform for natural language understanding.
\newblock \emph{arXiv preprint arXiv:1804.07461}.

\bibitem[{Wang et~al.(2023{\natexlab{a}})Wang, Liang, Meng, Sun, Shi, Li, Xu, Qu, and Zhou}]{wang2023chatgpt}
Jiaan Wang, Yunlong Liang, Fandong Meng, Zengkui Sun, Haoxiang Shi, Zhixu Li, Jinan Xu, Jianfeng Qu, and Jie Zhou. 2023{\natexlab{a}}.
\newblock Is chatgpt a good nlg evaluator? a preliminary study.
\newblock \emph{arXiv preprint arXiv:2303.04048}.

\bibitem[{Wang et~al.(2023{\natexlab{b}})Wang, Liang, Meng, Zou, Li, Qu, and Zhou}]{wang2023zero}
Jiaan Wang, Yunlong Liang, Fandong Meng, Beiqi Zou, Zhixu Li, Jianfeng Qu, and Jie Zhou. 2023{\natexlab{b}}.
\newblock Zero-shot cross-lingual summarization via large language models.
\newblock \emph{arXiv preprint arXiv:2302.14229}.

\bibitem[{Wang et~al.(2023{\natexlab{c}})Wang, Liang, Sun, Cao, and Xu}]{wang2023cross}
Jiaan Wang, Yunlong Liang, Zengkui Sun, Yuxuan Cao, and Jiarong Xu. 2023{\natexlab{c}}.
\newblock Cross-lingual knowledge editing in large language models.
\newblock \emph{arXiv preprint arXiv:2309.08952}.

\bibitem[{Wiseman et~al.(2017)Wiseman, Shieber, and Rush}]{wiseman2017challenges}
Sam Wiseman, Stuart~M Shieber, and Alexander~M Rush. 2017.
\newblock Challenges in data-to-document generation.
\newblock \emph{arXiv preprint arXiv:1707.08052}.

\bibitem[{Wu et~al.(2022)Wu, Zhang, and Li}]{wu2021text}
Xueqing Wu, Jiacheng Zhang, and Hang Li. 2022.
\newblock \href {https://doi.org/10.18653/v1/2022.acl-long.180} {Text-to-table: A new way of information extraction}.
\newblock In \emph{Proceedings of the 60th Annual Meeting of the Association for Computational Linguistics (Volume 1: Long Papers)}, pages 2518--2533, Dublin, Ireland. Association for Computational Linguistics.

\bibitem[{Xu et~al.(2017)Xu, Xu, Liang, Xie, Liang, Cui, and Xiao}]{xu2017cn}
Bo~Xu, Yong Xu, Jiaqing Liang, Chenhao Xie, Bin Liang, Wanyun Cui, and Yanghua Xiao. 2017.
\newblock Cn-dbpedia: A never-ending chinese knowledge extraction system.
\newblock In \emph{International Conference on Industrial, Engineering and Other Applications of Applied Intelligent Systems}, pages 428--438. Springer.

\bibitem[{Yang et~al.(2023)Yang, Xiao, Wang, Zhang, Bian, Yin, Lv, Pan, Wang, Yan et~al.}]{yang2023baichuan}
Aiyuan Yang, Bin Xiao, Bingning Wang, Borong Zhang, Ce~Bian, Chao Yin, Chenxu Lv, Da~Pan, Dian Wang, Dong Yan, et~al. 2023.
\newblock Baichuan 2: Open large-scale language models.
\newblock \emph{arXiv preprint arXiv:2309.10305}.

\bibitem[{Zhao et~al.(2023)Zhao, Zhou, Li, Tang, Wang, Hou, Min, Zhang, Zhang, Dong et~al.}]{zhao2023survey}
Wayne~Xin Zhao, Kun Zhou, Junyi Li, Tianyi Tang, Xiaolei Wang, Yupeng Hou, Yingqian Min, Beichen Zhang, Junjie Zhang, Zican Dong, et~al. 2023.
\newblock A survey of large language models.
\newblock \emph{arXiv preprint arXiv:2303.18223}.

\bibitem[{Zhong et~al.(2023)Zhong, Ding, Liu, Du, and Tao}]{zhong2023can}
Qihuang Zhong, Liang Ding, Juhua Liu, Bo~Du, and Dacheng Tao. 2023.
\newblock Can chatgpt understand too? a comparative study on chatgpt and fine-tuned bert.
\newblock \emph{arXiv preprint arXiv:2302.10198}.

\end{thebibliography}
\bibliographystyle{acl_natbib}


\end{document}